# Entity-Augmented Neuroscience Knowledge Retrieval Using Ontology and Semantic Understanding Capability of LLM

Pralaypati Ta[a,b], Sriram Venkatesaperumal[a], Keerthi Ram[a], and Mohanasankar Sivaprakasam[a,b]

[a] Sudha Gopalakrishnan Brain Centre, Indian Institute of Technology, Madras, Chennai, India.
[b] Department of Electrical Engineering, Indian Institute of Technology, Madras, Chennai, India.

Corresponding author: Pralaypati Ta (e-mail: pralay@gmail.com)

***Abstract*— Neuroscience research publications encompass a vast wealth of knowledge. Accurately retrieving existing information and discovering new insights from this extensive literature is essential for advancing the field. However, when knowledge is dispersed across multiple sources, current state-of-the-art retrieval methods often struggle to extract the necessary information. A knowledge graph (KG) can integrate and link knowledge from multiple sources. However, existing methods for constructing KGs in neuroscience often rely on labeled data and require domain expertise. Acquiring large-scale, labeled data for a specialized area like neuroscience presents significant challenges. This work proposes novel methods for constructing KG from unlabeled large-scale neuroscience research corpus utilizing large language models (LLM), neuroscience ontology, and text embeddings. We analyze the semantic relevance of neuroscience text segments identified by LLM for building the knowledge graph. We also introduce an entity-augmented information retrieval algorithm to extract knowledge from the KG. Several experiments were conducted to evaluate the proposed approaches. The results demonstrate that our methods significantly enhance knowledge discovery from the unlabeled neuroscience research corpus. The performance of the proposed entity and relation extraction method is comparable to the existing supervised method. It achieves an F1 score of 0.84 for entity extraction from the unlabeled data. The knowledge obtained from the KG improves answers to over 52% of neuroscience questions from the PubMedQA dataset and questions generated using selected neuroscience entities.***

## I. INTRODUCTION

The volume of published research articles in neuroscience has consistently increased over time [1] [2]. This extensive literature contains a wealth of knowledge in the form of new findings, claims, and counterclaims crucial for advancing neuroscience research and improving neurological patient care. However, the unstructured nature of this data makes extracting valuable insights from these articles complex and time-consuming [3]. More importantly, the interconnections of various published research works, especially their findings, remain largely unexplored as they have not been effectively linked [4]. Establishing such links is especially crucial for neuroscience. Fully depicting the findings on a given neuroscience topic would include uncovering the relationships between the various research streams that examine multiple facets of that topic [4] [5]. Hence, there is a need for an automated and practical solution to extract information from research articles and facilitate navigation between interrelated findings across related studies. Such a system would significantly enhance the knowledge discovery process [3] [5].

Dense passage retrieval, a specific technique under broader neural network-based retrieval methods, has emerged as a leading approach for extracting semantically similar knowledge from vast collections of textual data [6]. These methods utilize neural networks [8], like BERT [7], to represent questions and documents as dense vector embeddings within a common vector space, determining relevance through similarity scores. However, they often struggle to accurately retrieve all necessary passages for complex questions, particularly when the relevant information is dispersed across multiple documents with limited lexical overlap [9]. Knowledge graphs (KG) are effective for integrating knowledge from various sources, improving information retrieval, and enabling automated knowledge discovery. KGs explicitly store the relationships and address the challenges associated with dense passage retrieval by linking the research articles through shared entities [9].

The primary steps in converting unstructured content from neuroscience literature into a knowledge graph (KG) involve extracting and establishing relationships between neuroscience entities. Various approaches for extracting entities from unstructured text have been proposed, including rule-based, dictionary-based, and deep learning methods [15]. Recent studies have demonstrated that deep learning-based methods consistently outperform other techniques in neuroscience entity recognition tasks [12] [13] [14]. However, the deep-learning approaches require substantial amounts of annotated text to train their models [15], which can be expensive and time-consuming, especially in specialized fields like neuroscience. Hence, there has been growing interest in exploring zero or few-shot entity and relation extraction methods. Large Language Models (LLMs) have been studied for biomedical entity and relation extraction using zero or few-shot learning, and minimal instruction tuning [16] [17] [21-23]. But, LLMs often demonstrate lower recall, although high precision, for the biomedical entity extraction tasks [43]. Also, any effective question-answering system designed for a specialized domain like neuroscience must provide answers supported by research evidence. LLMs can answer domain-specific questions based on their parametric knowledge. However, when asked to provide proof for an answer, LLMs suffer from reference fabrication problems [27]. Recent studies have proposed leveraging knowledge graphs to overcome this challenge, utilizing a relevant subgraph to provide context for the LLM in generating answers [29].

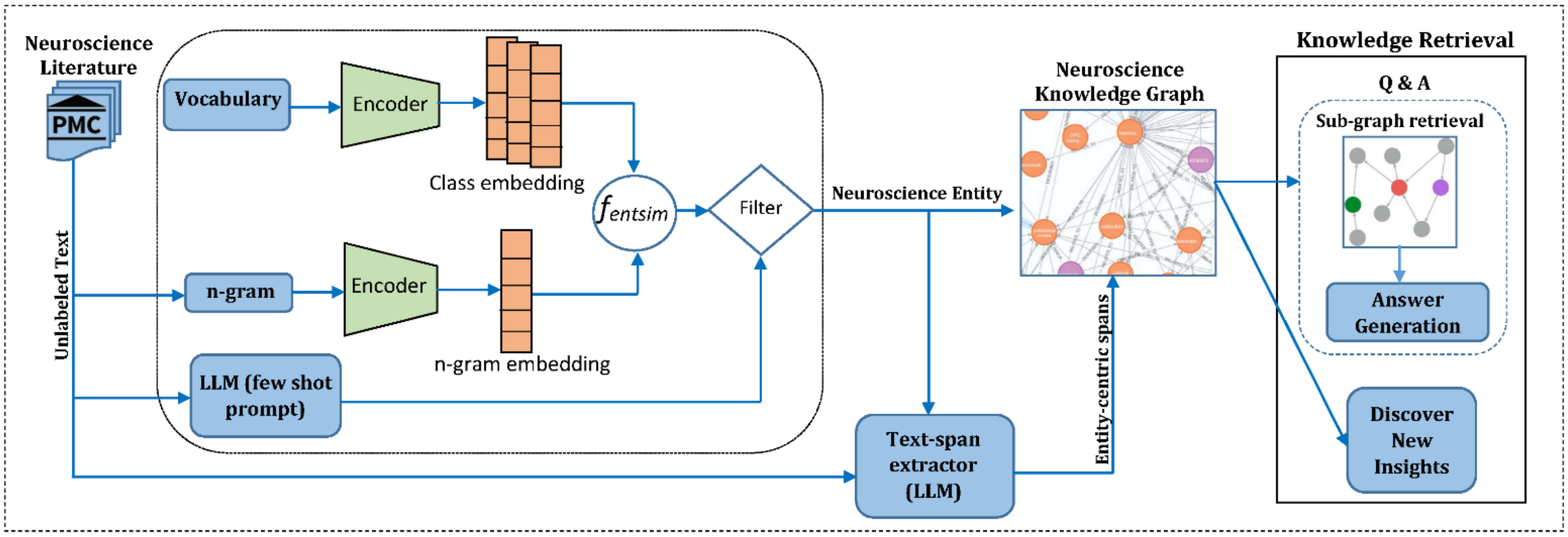

Fig. 1. Construction of neuroscience KG from unlabeled data and knowledge discovery using KG.

In this work, we construct a knowledge graph from an unlabeled neuroscience research corpus using zero-shot and few-shot methods, as depicted in Fig. 1. We propose a novel approach for neuroscience entity extraction by combining entity linking and LLM-based methods. First, we formulate entity extraction as an entity linking (EL) problem [19] that jointly performs entity mention detection and disambiguation. It uses Neuroscience Information Framework Standardized Ontologies (NIFSTD) [18] and semantic similarity to perform entity linking. The entities generated by this method are combined with those extracted by LLM, and the combined set is finally filtered to remove false positives. We evaluate the performance of our approach against the performance of various open-source language models, other similarity-based unsupervised approaches, and existing supervised approaches. Our approach shows strong performance, both in precision and recall, for the entity extraction task from unlabeled neuroscience texts.

We use an entity-augmented text extraction prompt to instruct LLM in extracting relevant text spans that express the relationships between entities in unlabeled neuroscience texts. To investigate the capability of LLM in extracting such text spans, we formulate a dense retrieval task using neuroscience-specific questions from the PubMedQA [25] dataset. Given a question, the task is to identify the corresponding context. We also evaluate the proposed relationship extraction approach using the publicly available WhiteText [64] dataset. Our evaluation shows that the entity-centric text spans extracted by LLM hold semantic relevance and describes the relations between the entities.

We construct a prototype knowledge graph from the 3248 neuroscience papers, 1764 full text & 1484 abstracts, employing the entity extraction and LLM-based relation extraction steps mentioned above (Fig. 1). We develop an entity-augmented subgraph retrieval module to extract pertinent subgraphs for a given question. An LLM can generate evidence-based answer to the question using the extracted subgraph as augmented knowledge. The efficacy of the extracted subgraph was assessed using an LLM as a judge [30] and neuroscience questions. Our experiment shows that the knowledge graph-based retrieval system performs better in retrieving relevant knowledge for a significant percentage of questions than the baseline dense passage retrieval [6] method. We investigate the LLMs' reference fabrication problems in the neuroscience domain and find that the issue also exists in LLMs while answering neuroscience questions. However, this issue is mitigated when the extracted subgraph enhances LLM's knowledge, as it includes valid references to the research articles used to construct it.

To summarize, our contributions are as follows:

- We present a novel approach for extracting neuroscience entities from unlabeled texts, combining entity linking and LLM-based extraction methods. Our approach achieves an F1 score of 0.84 and outperforms few-shot LLM-based and other techniques.
- We investigate LLMs' entity-specific text extraction capability, which is used to extract relationships between the entities

in neuroscience literature.

- We construct a neuroscience knowledge graph based on the entity and relation extraction techniques mentioned above. We develop an entity-augmented information retrieval module to extract the relevant sub-graph from this knowledge graph. Experiments conducted using neuroscience question indicate that leveraging the knowledge graph allows for improved answers to various questions, mainly when the information is dispersed across multiple documents.
- We show the reference fabrication issue prevalent in LLM's answers when addressing queries within the neuroscience domain, and the issue can be mitigated by using the relevant knowledge retrieved from KG.

## II. Related Work

Extracting structured information from unstructured text is essential for knowledge discovery. Many research studies have focused on extracting structured data from unstructured scientific texts in the biomedical domain [10] [20], including neuroscience [3] [11-14] [32], and also in other fields of science [22] [54]. The two most essential tasks in structured information extraction from the scientific literature are extracting domain-specific concepts, called entity, and their relations.

### A. Entity and Relation Extraction

The existing research works for entity and relation extraction in the biomedical domain can be divided into the following categories [45]: rule-based, ontology, and dictionary-based, using machine learning and deep learning techniques, along with the use of large language models in more recent studies. In dictionary-based approaches, the Unified Medical Language System (UMLS) is most commonly used [10] [11] [12]. QuickUMLS [31], MaxMatcher [62], and DeepLife [40] use approximate string matching to match the unstructured text with the terms in UMLS to extract medical concepts. Ontology has been used for extracting structured information related to the treatment of spinal cord injury [32]. Conditional Random Field (CRF) is the most commonly used machine learning model for extracting entities from unstructured text [12].

In DL-based methods, many works use the Bidirectional Long Term Short Term Memory Networks (Bi-LSTM) model combined with CRF for entity extraction tasks [12] [39]. In BioBERT-CRF [13], the output of the BioBERT layer is used as input to the CRF layer to extract the brain region entities. It achieved an F1 score better than Bi-LSTM-CRF. BioSEPBERT [14] connects a position pointer module to the output of BERT to extract brain regions and connectivity relations among them. Word sense disambiguation (WSD) is a task related to entity recognition and relation extraction. Various approaches using the embedding vector of the words and the corresponding concepts from the knowledge base have been explored in [42] for biomedical WSD. A comparison between dictionary-based, CRF-based, and DL-based methods for entity recognition was made in [12].

With the advent of LLMs, researchers are utilizing them for different types of NLP tasks in the biomedical domain [51]. BioNER-LLaMA [17] has fine-tuned LLAMA-7B models using instruction tuning for biomedical entity recognition. It has achieved performance comparable to the specialized PubMedBERT model. Zero-shot and few-shot learning approaches in LLMs have been explored for biomedical entity recognition and relation extraction [16] [21]. As shown in [16], the few-shot approach shows significant improvement but is still behind the specialized models such as PubMedBERT. In [47], in-context learning (ICL) in LLM is used to improve the accuracy of the information extracted from the clinical data. Using prompt engineering in GPT-4, [53] achieved an F1 score of 0.887 in biological entity recognition. In addition to the biomedical domain, LLMs are also used for information extraction from unstructured scientific texts of other domains [22] [54].

An entity may have many different surface forms [19]. Entity linking (EL) maps different non-standard surface forms of an entity to a standard label in the knowledge base [36]. The EL task includes two steps: surface form identification, referred to as mention detection (MD), and entity disambiguation (ED). The neural entity linking models use encoders to generate vector representations of entity mentions and candidate entities. Then, a ranking model is used to determine the matching score [56]. MedLinker [44] employs BiLSTM and CRF for embedding generation, utilizing cosine similarity to calculate the matching score. An end-to-end entity linking method is proposed in [55] to perform MD and ED jointly. It uses bidirectional LSTM to generate the embedding of entity mentions. In [46] and [48], the candidate entities are retrieved using an approximate nearest neighbor search. EL methods using n-gram are also proposed in the literature [52] [57].

Many studies have used entity-augmented or entity-focused retrieval to improve the performance in the information extraction task. CLEAR [24] utilizes entities from the input query to retrieve text from clinical notes, achieving superior performance compared to the baseline RAG [28]. EnFoRe [34] identifies entities from the query and computes their importance score given the query and passages. These scores are then used to determine the query-passage relevancy. An entity-centric information retrieval approach is proposed in [37] in which the bridge entities connect the relevant paragraphs in multi-hop question answering.

Extracting the most relevant sentences or spans of text from the unstructured text is crucial in many NLP tasks. The Machine Reading Comprehension (MRC) task involves understanding unstructured text and utilizing the applicable span of tokens to answer questions [36]. The work in [33] combines term-based and neural semantic retrieval to extract a set of sentences to answer the query. In [50], the NER task is formulated as an MRC task that locates and extracts the relevant span. Clinical concepts and relation extraction as an MRC task have been explored in [58]. It uses BERT and LLM with prompt-based question answering to extract the pertinent text spans.

### B. Biomedical Knowledge Graph

Domain-specific knowledge graph (KG) stores verified domain knowledge in a structured form. Researchers have created several knowledge graphs for the biomedical domain [39]. KGHC [10] is a knowledge graph for hepatocellular carcinoma created from structured and unstructured data. PrimeKG [41] is a multimodal knowledge graph constructed using 20 structured knowledge bases for precision medicine analysis. A KG for heart failure has been developed in [23] using LLM and prompt engineering. However, there are only a few works on knowledge graph construction in neuroscience. A knowledge graph development approach to discover knowledge from the Dementia research articles has been proposed in [3]. In the research work [11], a knowledge graph from unstructured research abstracts related to autism spectrum disorder has been constructed with minimum supervision. One primary use of KG is question answering. Various methods have been suggested to extract the relevant sub-graph from the KG to answer questions [29] [35]. KGAREVION [38] uses KG to verify and filter LLM-generated triplets containing the answer to the input query.

We identified the following research gaps in KG construction for the neuroscience domain. First, there is limited work on neuroscience entity extraction methods that work on unlabeled data and can accurately extract any neuroscience entity. Second, LLM's semantic understanding capability in neuroscience research texts has not been explored much. Lastly, to our knowledge, limited research exists on constructing knowledge graphs related to neuroscience.

## III. Methods

Our proposed entity extraction method uses an entity-linking approach to extract the neuroscience entities. It links the entity names from ontologies to the text spans of unlabeled texts, which helps to improve the recall. Entities from the entity-linking step are combined with those extracted by LLM, which has high precision, to enhance the overall accuracy. The details of the formulation and methodology for the entity extraction task are described in subsections III-A and III-B.

We used LLM to extract entity-centric text spans that describe the relation among entities. The semantic relevance of the spans has been evaluated using a proposed dense retrieval task. We created a KG using the entities and the corresponding text spans. The KG and the text spans help generate evidence-based and improved answers to neuroscience questions. The formulation of the dense retrieval task and the methodology of KG construction are discussed in subsections III-C & III-D.

### A. Extraction of Neuroscience Entities

Given an unstructured text document $D = \{w_1, w_2, \ldots, w_n\}$ , where $D$ is a sequence of words $w_i$, and a vocabulary of entities $V = \{e_k\}_{k \in [1, |V|]}$ from the knowledge base, the entity extraction task, formulated as an entity-linking problem [19] [48], generates a list of $\{(w_{ij}, e_k)\}$ pairs [55], where $w_{ij}$ is a sequence of words from $D$ and $e_k \in V$.

We developed the vocabulary $V$ of neuroscience entities by recursively traversing the NIFSTD [18] ontology tree, including the imported ontologies. We collected each class's labels, preferred and alternate labels, synonyms, and abbreviations whenever possible [12]. To further enrich the neuroscience vocabulary, we incorporated ontology classes from the "*Nervous System*" and "*Nervous System Diseases*" subtrees of the Medical Subject Headings (MeSH) ontology. The generated vocabulary underwent a review and cleaning process; some common English words were removed using the Python library NLTK. Finally, the vocabulary has ~400K neuroscience entity strings representing different surface forms of the entities.

The n-grams of a text document $D = \{w_1, w_2, \ldots, w_n\}$ are the sequence of words $s_{ij} = \{w_i, \ldots, w_j\} \forall i, j$ where $\{i \in \{1, .., |D| - n + 1\},\ j = (i + n - 1)$ and $n \in \{1, .., \max(n)\}$. We generated all possible n-grams [52] [57] from a text document $D$ for a specific maximum value of n and compared them against the entities in $V$ using a similarity function [11] *entsim* to generate the list of $\{(w_{ij}, e_k)\}$ pairs. The similarity function returns the following set:

$$E_{all} = \{(w_{ij}, e_k) \mid entsim(w_{ij}, e_k) \geq \alpha\} \tag{1}$$

where $\alpha \in [0, 1]$ is similarity threshold. The entities from set $E_{all}$ that match the longest non-overlapping word sequences are ultimately considered as the entities present in document $D$.

We explored two similarity functions: 1) fuzzy matching utilizing Levenshtein Distance and 2) cosine similarity through the semantic retrieval model [19] [56]. The fuzzy matching-based similarity check was implemented using the Python library theFuzz. We employed two models for cosine similarity with the semantic retrieval model: NV-Embed-v2 [26] and a fine-tuned BAAI General Embedding (BGE) large model [59]. NV-Embed-v2 is an LLM-based embedding model, and it was one of the top-scored models in the embedding leaderboard [61] for retrieval tasks while we performed the experiments. For fine-tuning the BGE model, we generated two different strings for each entity name in the vocabulary, having a maximum of two edit distances from the actual names collected from the ontology [40]. We first conducted continual pre-training of the BGE model using all entity names, including actual names and their edit distance variants, to adapt the model specifically to neuroscience entity names. During fine-tuning, the modified names for an entity were treated as queries, while the corresponding actual name served as a positive sample for the retrieval task. Five negative samples were also hard-mined for each modified entity name.

We conducted experiments using four open-source language models: Llama-3.1-70B, Llama-3.3-70B, Llama-3.1-405B [65], and BioMistral-7B [66], focusing on extracting neuroscience entities from unstructured research articles. The three Llama models are general-purpose models, whereas BioMistral-7B is specifically tailored for the biomedical field. We employed a one-shot prompt for this task, and the models generated outputs based on their parametric knowledge. Given an unstructured document $D$, the output of entity extraction task using LLMs can be formulated as follows:

$$f_{ex}(D) = \left\{ t \mid t \in D_{span}, e \in \mathcal{E}, P(t, e \mid D) \geq \tau \right\} \quad (2)$$

where $D_{span}$ denotes the text spans in document $D$, $\mathcal{E}$ represents the set of neuroscience entities present in the parametric knowledge of the model, $P(t, e \mid D)$ is the joint probability of span $t$ being a neuroscience entity $e$, $\tau$ is a threshold. Since $t$ is a span of $D$, it must be mapped to a standard form of an entity, which is discussed in the entity disambiguation step below.

We found that LLM-based entity extraction methods achieve higher precision but lower recall. In contrast, the similarity-based retrieval using NV-Embed-v2 demonstrates higher recall but relatively lower precision. To benefit from both these methods, we developed a new approach combining the entities extracted by the LLM-based and NV-Embed-v2 similarity methods. We then filter this combined set again using a one-shot prompt and Llama-3.3-70B model to eliminate false positives. This method was finally used to extract entities in the subsequent steps. The implementation details of this method, the cosine similarity threshold, and prompts used for entity extraction and filtering can be found in the *Supplementary material, Sec. H*.

## B. Entity Disambiguation

A neuroscience entity can appear in various surface forms, including synonyms, abbreviations, and lexical variations such as singular and plural or hyphenated versions. It is essential to map these different representations to a standard name before using them in downstream tasks like knowledge graph (KG) construction or KG-based retrieval. When creating the vocabulary $V$ of neuroscience entities from the NIFSTD [18] and MESH ontologies, we use the *skos:prefLabel*, if available, as the standard name for each entity; otherwise, *rdfs:label* is used. Additionally, we maintain a mapping table to map other names and abbreviations to the standard name. For entity names extracted by LLM-based methods, the cosine similarity with a high threshold value is used to map the document spans to standard names in vocabulary [24]; if no similar name is found, the span is treated as a new entity and added to the vocabulary.

## C. Evaluation of Entity Centric Text Span Retrieval

We utilized PubMedQA [25] labeled and unlabeled QA instances to assess the entity-centric retrieval capabilities of large language models (LLMs). The PubMedQA dataset consists of question-context pairs derived from the abstracts of research articles. To focus specifically on neuroscience, we filtered the neuroscience question-context pairs from the PubMedQA dataset using the Llama-3.3-70B model in two steps. First, we provided a one-shot prompt to the Llama-3.3-70B model, asking it to identify and filter the neuroscience-specific contexts. Then, we prompted the model again to extract the neuroscience entities from the filtered contexts. Ultimately, we selected the contexts that contained at least one neuroscience entity for the subsequent tasks. Using these neuroscience-specific question-context pairs, we formulated a retrieval task in which, given a question, the corresponding context needs to be retrieved.

We used the Llama-3.3-70B model with an entity-augmented text extraction prompt to extract entity-centric texts from the contexts. The model is given a set of entities mentioned in a context and asked to summarize the entity-specific texts from the context paragraph [24]. We then used the NV-Embed-V2 model to generate the embedding for the questions, contexts, and entity-centric text spans.

We developed an algorithm for the retrieval task, as illustrated in "Algorithm 1," which utilizes a weighted combination of similarity scores of complete context text and entity-centric text spans. The embedding of the entire context captures the overall semantics, while the embedding of entity-specific spans focuses on the semantics relevant to particular entities. When two contexts closely match a given question, we calculate the overall similarity score by incorporating the similarity scores of the entity-specific texts, thereby enhancing retrieval performance. We conducted experiments to determine how varying the weight of text span similarity affects the overall similarity score calculation. Our findings revealed that the performance of the retrieval task improves when the weight of the entity-specific text span similarity score is gradually increased, particularly as the difference in the full context similarity score grows. The final weight assigned to the entity-specific text spans is presented in Table 1.

We also used the Llama-3.3-70B model with entity-augmented text extraction prompt to extract the relevant text span describing the relation between the brain regions in the WhiteText [64] dataset. The WhiteText dataset is created from 2817 abstracts containing 20283 annotated brain regions and 5208 connectivity relations between the regions. Given a text and a pair of brain regions from the WhiteText dataset, the model was asked to extract the relevant text describing the relation between the regions.

**Algorithm 1** Context Retrieval Using Entity Specific Spans

$v_q$: Normalized query embedding vector
$C$: All normalized context embedding vectors
$S_c$: All normalized embedding vectors of entity-specific spans of context $C$
$threshold$ : Default value 0.05

**get_similar_context ($q$)** // $q$ is the query
$S_1, S_2 \leftarrow top2values(v_q \cdot C^T)$ // $S_1, S_2$ are similarity scores
$C_1 \leftarrow$ Context corresponding to similarity score $S_1$
$C_2 \leftarrow$ Context corresponding to similarity score $S_2$
score_diff $\leftarrow S_1 - S_2$ // $S_1 > S_2$
if score_diff $< threshold$
$ES_1 \leftarrow \max(v_q \cdot S_{c1}^T)$
$ES_2 \leftarrow \max(v_q \cdot S_{c2}^T)$
$span_weight \leftarrow$ get_span_weight (score_diff)
$S_1' \leftarrow S_1 * (1 - span_weight) + ES_1 * span_weight$
$S_2' \leftarrow S_2 * (1 - span_weight) + ES_2 * span_weight$
$S_1 \leftarrow \max(S_1, S_1')$
$S_2 \leftarrow \max(S_2, S_2')$
if $S_1 > S_2$
**return** $C_1$
else
**return** $C_2$

TABLE I
WEIGHTAGE OF SPAN SIMILARITY SCORE

| Score difference | Span weightage |
|---|---|
| <= 0.01 | 0.10 |
| <= 0.02 | 0.15 |
| <= 0.03 | 0.20 |
| <= 0.04 | 0.25 |
| <= 0.05 | 0.30 |

### D. KG Construction and Knowledge Retrieval Using KG

We used Entrez Programming Utilities to extract research articles from the PubMed Central (PMC) archive. The PMC archive was searched using various subcategories of the *Neurosciences* MeSH category. However, this search yielded limited full-text neuroscience publications, so we collected several hundred additional high-quality articles from reputable neuroscience journals. More detailed information about the selection process and the selected articles can be found in the *Supplementary material, Sec. A & B*. The above process yielded 1,764 full-text articles and 1,484 abstract-only research articles, which were utilized to construct the prototype knowledge graph. The texts of these research articles were downloaded and organized into different paragraphs corresponding to the various sections of each article.

We extracted 22547 neuroscience entities from these research articles. The relationships among the entities were identified using the Llama-3.3-70B model through the entity-augmented text extraction prompt. The model was given a paragraph along with a set of entities from that paragraph and was asked to extract all relevant text spans that express the relationships between entities [58]. We also extracted the relevant text spans that describe the relation between an entity and the containing paragraph. The constructed knowledge graph contains only two generic edge types: RELATED_TO and DESCRIBES. The extracted text spans, which convey the relationships in natural language, are stored as edge attributes, providing additional context for the edges. In total, we extracted 612272 relations. The constructed knowledge graph can be defined as $G = (V, E)$, where $V = V_e \cup V_p$ and $E = E_{ee} \cup E_{ep}$, such that

- $V_e = \{v_e\}$, where $v_e = (id, entity_name)$ represent a node for a neuroscience entity.
- $V_p = \{v_p\}$, where $v_p = (id, paragraph_path)$ represent a node for a paragraph from research articles.
- $E_{ee} = \{e_{ee}\} \subseteq V_e \times V_e$, where $e_{ee} = (id,\ relation_text,\ paragraph_path)$ represents an entity-entity relationships "RELATED_TO".
- $E_{ep} = \{e_{ep}\} \subseteq V_e \times V_p$, where $e_{ep} = (id,\ relation_text)$ represents an entity-paragraph relationships "DESCRIBES".

We developed an entity-augmented retrieval algorithm [24], as outlined in "Algorithm 2" to extract the subgraph relevant to a given question. The algorithm returns a set of paragraphs that can be used to generate the answer to the question using an LLM. To evaluate the effectiveness of the subgraph retrieval algorithm, we used the neuroscience-specific questions from the PubMedQA [25] dataset. Per our analysis mentioned in the previous section, PubMedQA labeled and unlabeled datasets contain 8579 neuroscience-specific questions. However, because of the limited number of articles in the KG, "Algorithm 2" can retrieve relevant texts only for 460 questions. Based on our analysis, none of the neuroscience research articles from which these 460 PubMedQA questions were extracted are present in the KG. We evaluated the relevance and quality of these texts retrieved from KG to answer these questions against the baseline dense passage retrieval. We also investigated the knowledge retrieval capability when the knowledge is dispersed in multiple documents, using synthetically generated 493 neuroscience questions. More details about the method for generating the synthetic questions can be found in the *Supplementary material, Sec. C.*

To compare the paragraphs retrieved by our algorithm against those from the baseline dense passage retrieval, we retrieved another set of paragraphs by comparing the cosine similarity of the paragraph and query embedding vectors generated by the NV-Embed-v2 model. The Gemini 2.5 Flash model was then tasked with determining which set of paragraphs is best suited to answer the question. To mitigate the position bias [30], we asked the model to evaluate the sets of paragraphs twice by switching their positions in the prompt. Additionally, we ensured both sets contained the same number of paragraphs to minimize verbosity bias [30].

**Algorithm 2** Entity Augmented Sub-graph Retrieval

$v_q$: Normalized query embedding vector

**get_relevant_paragraphs ($q$)** // $q$ is the query

Extract the neuroscience entities $Ent_q$ from $q$, where

$Ent_q = \{ ent_q | \, ent_q \; is \; a \; neuroscience \; entity \; in \; q\}$

$V_{ent} = \{v_{ent} \mid v_{ent} \subseteq V_e \; and \; entity_name(v_{ent}) \in Ent_q \}$

if $|V_{ent}| == 1$

$V_{ent}$ has one vertex $v_{ent}$

Collect all relevant edge and paragraph node pairs $P_{ep}$ where,

$P_{ep} = \{(e_{ep}, v_p) \mid e_{ep} = (v_{ent} \times v_p) \in E_{ep} \; and \; v_p \in V_p$

Collect the relation descriptions and paragraph texts $T_{ep}$ where

$T_{ep} = \{(relation_text(e_{ep}), para_text(v_p) \mid (e_{ep}, v_p) \in P_{ep}\}$

else if $|V_{ent}| == 2$

$V_{ent}$ has two vertices $v_{ent}^1$ and $v_{ent}^2$

Collect all the edges $E' \subseteq E_{ee}$ in the paths between $v_{ent}^1$ and $v_{ent}^2$ where max path length = 2

Collect the relation descriptions and paragraph texts $T_{ep}$ where

$T_{ep} = \{(relation_text(e_{ee}), para_text(e_{ee}) \mid e_{ee} \in E'\}$

else

Extract two most important entities from $q$ using LLM

$v_{ent}^1$ , $v_{ent}^2$ ← vertices corresponding to the important entities

$v_{ent}^{other}$ ← vertices corresponding to the other entities

Collect all the edges $E_1' \subseteq E_{ee}$ in the paths between $v_{ent}^1$ and $v_{ent}^{other}$ where max path length = 2

Collect all the edges $E_2' \subseteq E_{ee}$ in the paths between $v_{ent}^{other}$ and $v_{ent}^2$ where max path length = 2

Collect the relation descriptions and paragraph texts $T_{ep}$ where

$T_{ep} = \{(relation_text(e_{ee}), para_text(e_{ee}) \mid e_{ee} \in E_1' \cup E_2'\}$

Re-rank the texts in $T_{ep}$ using semantic similarity score against $v_q$

$top_k$ ← top k paragraphs according to rank

**return** $top_k$

## *E. Analysis of the Reference Fabrication Issue in LLMs*

We generated 10 neuroscience questions to investigate the issue of reference fabrication in large language models (LLMs). Three LLMs, Llama-3.1-405B, Llama-3.3-70B, and gemini-2.0-flash-thinking-exp-01-21, were asked to answer these questions and provide the corresponding reference publications. Each model was required to include the title and DOI of the referenced publications.

Llama-3.1-405B cited 40 references in its answers, while Llama-3.3-70B and gemini-2.0-flash-thinking-exp-01-21 used 38 and 41 references, respectively. To verify the authenticity of these references, we searched for the publications using their DOIs and titles (for the closest title match) in the Semantic Scholar database, utilizing the Semantic Scholar Academic Graph API [60]. If the search returned any result, we manually compared the titles to verify whether the cited reference exists and relevant.

TABLE II
PERFORMANCE OF ENTITY EXTRACTION APPROACHES (MACRO-AVERAGED VALUES FOR RELAXED MATCH)

| Text | Approach | Model | Precision | Recall | $F_1$ |
|---|---|---|---|---|---|
| Neuroscience Research Articles | One-shot prompt | Llama-3.1-405B | **0.973** | 0.630 | 0.747 |
| | | Llama-3.1-70B | 0.917 | 0.498 | 0.625 |
| | | Llama-3.3-70B | 0.950 | 0.587 | 0.708 |
| | | BioMistral-7B | 0.772 | 0.412 | 0.517 |
| | Entity Linking Using Cosine similarity | NV-Embed-v2 | 0.612 | 0.656 | 0.627 |
| | | Finetuned bge-large-en-v1.5 | 0.824 | 0.296 | 0.417 |
| | Entity Linking Using Fuzzy similarity | ---- | 0.855 | 0.476 | 0.596 |
| | One-shot prompt + Entity Linking + Filtering | Llama-3.3-70B, NV-Embed-v2 | 0.897 | **0.806** | **0.842** |
| | | BioMistral-7B, NV-mbed-v2 | 0.573 | 0.457 | 0.476 |
| Clinical Text | One-shot prompt + Entity Linking + Filtering | Llama-3.3-70B, NV-Embed-v2 | 0.710 | 0.789 | 0.746 |

TABLE III
PERFORMANCE OF ENTITY EXTRACTION FOR WHITETEXT DATASET

| Extraction Method | NER (Lenient) Recall(%) | RE Recall(%) |
|---|---|---|
| BioSEPBERT [14] | 93.29 | **88.47** |
| BioSEPBERT-DEN [14] | **94.59** | --- |
| Ours | 93.05 | 84.16 |

## IV. RESULTS AND DISCUSSION

### A. Extraction of Neuroscience Entities

The proposed entity extraction method can be used to identify neuroscience entities from any document type, including neuroscience research articles and unstructured clinical notes. To evaluate this method, we selected 12 neuroscience research papers that cover a diverse range of topics, details of which can be found in *Supplementary Table S1*. Entities were extracted from the abstracts of these articles using different extraction techniques, resulting in a total of 1,209 neuroscience entities. We manually reviewed the entities extracted from each abstract and checked for correct, incorrect, and missed entities. The performance of different unsupervised entity extraction methods is shown in Table II. The scores were calculated using relaxed-match criteria in which the type of extracted entity matches the mentioned entity, and the spans overlap.

The LLM-based methods have high precision but lower recall. Also, LLM-based methods extract the exact text spans from the given text, which often has different surface forms, e.g., "*peripheral neuron*" and "*peripheral neurons*". It's necessary to map them to the standard form before any downstream task. As expected, the performance improves among LLMs as the number of parameters of the model increases. However, the hardware cost of running the models also rises with the increased parameter count. BioMistral-7B is a biomedical domain-specific model having one-tenth the parameters and running cost of the Llama-3.1-70B or Llama-3.3-70B model. However, performance degradation is not that high, highlighting the advantage of domain-specific training. We ran Llama-3.1-70B and Llama-3.3-70B models using 4 A100 GPUs on an on-premises NVIDIA DGX server. For the Llama-3.1-405B model, we utilized the NVIDIA NIM APIs. More details about the generation hyperparameters for LLMs can be found in the *Supplementary material, Sec. G*.

The entity extraction methods that utilize neuroscience vocabulary for entity linking require less expensive hardware. These methods map the span of the source texts to specific vocabulary terms. As a result, the generated entities exhibit fewer lexical variations; however, these methods tend to have lower precision than those based on large language models (LLMs). They also show lower recall rates, except for the entity linking using the NV-Embed-v2 model.

Entity linking with NV-Embed-v2 has the highest recall. As a result, the combined approach of the one-shot method using Llama-3.3-70B and the extraction method based on NV-Embed-v2 shows a significantly higher F1 score of 0.84 than the others. Combining the NV-Embed-v2 method with Llama-3.1-405B or larger models could further enhance performance, though this would also increase costs for large-scale entity extraction.

The proposed method can also be employed to extract neuroscience entities from unlabeled and unstructured clinical texts. We selected discharge reports of 10 patients having different nervous system diseases from the MIMIC-IV [63] dataset to evaluate the proposed method for clinical texts. The details of the selected ICD codes can be found in *Supplementary Table S2*. The performance of the proposed entity extraction method for the discharge reports is shown in Table II. It achieved an F1 score of 0.746 for clinical texts, which is lower than that of the research articles. Entity extraction from clinical notes is more challenging than entity extraction from research articles, as clinical notes include diverse information besides the neurology-related texts, such as family history, medication details, diagnostic and pathology test results. The texts in clinical notes are less formal than in research articles. It often contains incomplete sentences due to de-identification and sometimes spelling mistakes. Also, abbreviations are extensively used in clinical texts.

We also evaluated the performance of the proposed entity extraction method for the WhiteText [64] dataset and compared it with the SOTA model BioSEPBERT [14]. As reported in BioSEPBERT [14], we calculated the scores using lenient matching, in which the predicted boundaries overlap with the annotated boundaries. We only report the Recall score, i.e., the percentage of brain regions correctly identified by our method, in Table III for comparison. By design, our method extracts other entities besides the brain regions, and those are not false positives; hence, precision cannot be compared. Our entity extraction method achieved a recall of 93.05%, comparable to that of BioSEPBERT, without using any labeled data.

TABLE IV
PERFORMANCE OF PUBMEDQA CONTEXT RETRIEVAL TASK

| Threshold | Question Type | No. of Questions | Span Weightage | Average Precision@1 |
|---|---|---|---|---|
| 0.05 | All | 8579 | 0 | 93.43 |
| | | | 1 | 93.15 |
| | | | Variable | **93.85** |
| | Top 2 context similarity < 0.05 | 1358 | 0 | 62.96 |
| | | | 1 | 61.19 |
| | | | Variable | **65.61** |
| 0.01 | All | 8579 | 0 | 93.43 |
| | | | 1 | 93.50 |
| | | | Variable | **93.61** |
| | Top 2 context similarity < 0.01 | 312 | 0 | 43.59 |
| | | | 1 | 45.51 |
| | | | Variable | **48.72** |

## B. Entity Centric Retrieval & Relation Extraction

The results of the PubMedQA retrieval task with different weights for entity-centric spans are shown in Table IV. This table highlights how the entity-centric spans and their weightage in the similarity calculation impact the accuracy of the retrieval process. The retrieval task involving the PubMedQA dataset is challenging because the context section containing the actual answer to the query has been removed. When span weightage is zero, retrieval is performed solely based on the embedding vectors generated by NV-Embed-V2, achieving an impressive accuracy of 93.43%. Improving this accuracy further is even more challenging. However, as we can see in Table IV, the overall accuracy is further enhanced to 93.85% by using the varying weightage of the entity-centric spans in the similarity score calculation.

Notably, the enhancement in accuracy is even more pronounced when we focus on questions where the top two retrieved contexts exhibit a minimal difference in their similarity scores. In these cases, the accuracy improves by 2.65% and 5.13% when the differences are less than 0.05 and 0.01, respectively. This increase indicates that the entity-centric spans extracted by LLM hold significant semantic relevance. If the entity-centric spans are given a maximum weight of 1 in the overall similarity calculation, the retrieval accuracy decreases to 93.15%, emphasizing the importance of the complete text's semantics. Accuracy improves only when the final similarity calculation combines context-level and entity-centric semantics.

The performance of the relation extraction task for the WhiteText [64] dataset is shown in Table III. Our proposed method extracts all possible types of relations, besides the connectivity, between the mentioned brain regions. We only report the Recall score in Table III for comparison with BioSEPBERT [14]. A total of 17253 text spans were extracted describing the relation between the regions. Of these, 4383 match the connectivity relations mentioned in the WhiteText dataset, achieving a recall of 84.16%.

## C. Knowledge Retrieval Using KG

The knowledge graph contains valuable information from the research articles and can provide new insights, such as identifying the most and least explored research areas. Table V presents some of the most and least mentioned neuroscience entities, determined by the degree of their corresponding nodes in the knowledge graph. The *Hippocampus* and *Astrocytes* are among the most frequently discussed entities. *Moebius syndrome* and *Stria medullaris* are the least mentioned, suggesting that these concepts have not been widely explored in the selected research articles.

TABLE V
MOST AND LEAST EXPLORED ENTITIES IN KG

| Neuroscience terms | Node degree |
|---|---|
| Hippocampus | 3002 |
| Astrocytes | 2595 |
| Amygdala | 2220 |
| Schizophrenia | 1436 |
| Pyramidal neuron | 656 |
| Oligodendrocyte Precursor Cell | 168 |
| 5-HT1A receptor | 61 |
| Stria medullaris | 1 |
| Moebius syndrome | 1 |

TABLE VI
COMPARISON OF THE ANSWERS GENERATED FROM KG

| Dataset | Question Type | Position of Sub-graph Retrieved Texts in Prompt | |
|---|---|---|---|
| | | First | Second |
| PubMedQA | Total Evaluated | 460 | 460 |
| | Sub-graph Retrieved Texts Deemed Best | 242 (52.61%) | 260 (56.52%) |
| | Best Texts Common in Both Positions | 219 | |
| Synthetically Generated Questions | Total Evaluated | 493 | 493 |
| | Sub-graph Retrieved Texts Deemed Best | 277 (56.19%) | 270 (54.77%) |
| | Best Texts Common in Both Positions | 202 | |

Table VI shows the relevance of the paragraphs retrieved by the entity-augmented sub-graph retrieval algorithm for answering the PubMedQA and synthetically generated neuroscience questions. When paragraphs retrieved by the proposed algorithm were presented first in the LLM prompt, the retrieved texts for 242 out of 460 (52.61%) PubMedQA questions were deemed the best. Conversely, when paragraphs were placed second, texts for 260 out of 460 (56.52%) PubMedQA questions were rated the best. Among the two sets of questions, 219 are common, for which the paragraphs retrieved using KG were identified as best. The knowledge graph includes fewer research articles, and our subgraph retrieval algorithm can answer a limited number of PubMedQA questions. However, the text retrieved from the knowledge graph is more suitable for answering over 50% of those questions. These results are promising. Even with a relatively small KG, the combination of knowledge graph-based retrieval and embedding-based semantic retrieval can significantly enhance knowledge discovery in the neuroscience field. We also asked the Gemini 2.5 Flash model to explain why it chose a specific set of paragraphs as the best for answering a question. For reporting purposes, a few such explanations have been mentioned in the *Supplementary material, Sec. D*.

For the generated questions, the retrieved texts for 277 out of 493 (56.19%) questions were deemed the best when paragraphs retrieved by our algorithm were presented first in the LLM prompt. Conversely, when paragraphs were placed second, texts for 270 out of 493 (54.77%) questions were rated the best, and 202 questions were common to both sets, confirming that the paragraphs retrieved using KG were superior for a significant percentage of the questions. Since the answers to these questions are dispersed across different articles or parts of articles, our analysis shows that the dense retrieval fails to retrieve all the relevant texts. We observed that the extracted relevant text span improves the retrieval performance and assists LLM in relating the important parts of the paragraph while generating the answers. A detailed analysis of the paragraphs retrieved and the generated answers for a few questions can be found in the *Supplementary material, Sec. E & F*.

TABLE VII
ANALYSIS OF LLM'S REFERENCE FABRICATION PROBLEM

| **Search By** | **Reference Type** | **gemini-2.0-flash-thinking-exp-01-21** | **Llama-3.1-405B** | **Llama-3.3-70B** |
|---|---|---|---|---|
| -- | Total Cited References | 41 | 40 | 38 |
| DOI | Not Found | 16 | 18 | 16 |
| | Title Mismatch | 25 | 22 | 22 |
| | Incorrect DOIs (%) | 41 (100%) | 40 (100%) | 38 (100%) |
| Title (Closest Match) | Not Found | 27 | 32 | 23 |
| | Title Mismatch | 11 | 3 | 1 |
| | Incorrect Titles (%) | 38 (92.68%) | 35 (87.50%) | 24 (63.16%) |

## *D. Reference Fabrication Issue in LLMs*

We analyzed the reference fabrication issue present in the answers generated by LLM for the neuroscience questions. The results of our study are summarized in Table VII. The DOIs are incorrect in all the references for all three models. Specifically, the DOIs were either fabricated and could not be located, or the titles of the references generated by the models did not match the titles of the actual publications associated with those DOIs.

When we searched for the references using their titles, we found that a significant percentage of them either could not be found or that the closest matched title was different. Among the three models, the gemini-2.0-flash-thinking-exp-01-21 model had the highest rate of incorrect titles at 92.68%, while Llama-3.3-70B had the lowest at 63.16%.

Our analysis shows that LLMs generate spurious neuroscience references when the answers are generated from their parametric knowledge. Also, we only checked for the existence of the references, but it still needs to be analyzed if the references are relevant to the generated answers. This analysis highlights the significance of knowledge graphs in providing evidence-based answers in specialized fields like neuroscience. The answers generated by LLM based on the sub-graph retrieved from KG contain valid and correct references. Examples of such answers are in the *Supplementary material, Sec. F.*

## V. Conclusion

In this work, we present our research contributions focused on methods for knowledge discovery from neuroscience research articles. We developed an unsupervised entity extraction method that leverages LLMs and entity linking with specialized neuroscience vocabulary. This method works without requiring annotated data and can accurately extract neuroscience entities from research articles. The proposed method is also generic and can be applied to other specialized domains, such as Cancer, for which a comprehensive ontology is available. Next, we explored the capability of LLMs to extract entity-centric relevant texts from neuroscience literature. Our findings indicate that the texts extracted by the LLM, which are focused on specific entities, possess semantic relevance. We utilized this text extraction capability to identify relationships between entities mentioned in neuroscience texts and subsequently constructed a knowledge graph.

We demonstrated the effectiveness of the knowledge graph by applying an entity-augmented information retrieval algorithm that answers neuroscience questions by revealing new knowledge. This KG was constructed from just a few thousand neuroscience research articles, while millions of articles have been published. More research articles should be added to enhance the KG's effectiveness in discovering neuroscience knowledge.